\documentclass[letterpaper, 10 pt, conference]{ieeeconf}  

\IEEEoverridecommandlockouts                              

\usepackage{makecell}
\usepackage[pagebackref,breaklinks,colorlinks]{hyperref}
\usepackage{amssymb}
\usepackage{float}
\usepackage{graphicx}
\usepackage{adjustbox}
\usepackage{multirow}
\usepackage{tabularx}
\usepackage[table,xcdraw]{xcolor}
\usepackage{amsmath}
\usepackage{listings}
\usepackage{xcolor}
\usepackage[utf8]{inputenc}
\usepackage[linesnumbered,ruled,vlined]{algorithm2e}
\usepackage{array}  
\usepackage{booktabs} 

\usepackage{pifont}
\newcommand{\xmark}{\ding{55}}%

\title{\LARGE \bf DriveLMM-o1: A Step-by-Step Reasoning Dataset and Large Multimodal Model for Driving Scenario Understanding}
\author{Ayesha Ishaq$^{1}$, Jean Lahoud$^{1}$, Ketan More$^{1}$, Omkar Thawakar$^{1}$, Ritesh Thawkar$^{1}$,\\ Dinura Dissanayake$^{1}$, Noor Ahsan$^{1}$, Yuhao Li$^{1}$, Fahad Shahbaz Khan$^{1,2}$,  Hisham Cholakkal$^{1}$, \\Ivan Laptev$^{1}$, Rao Muhammad Anwer$^{1}$, Salman Khan$^{1,3}$
\thanks{$^{1}$ Mohamed Bin Zayed University of Artificial Intelligence}
\thanks{$^{2}$ Linköping University, $^{3}$ Australian National University}
}
\begin{document}

\maketitle
\thispagestyle{empty}
\pagestyle{empty}


\begin{abstract} 
While large multimodal models (LMMs) have demonstrated strong performance across various Visual Question Answering (VQA) tasks, certain challenges require complex multi-step reasoning to reach accurate answers. 
One particularly challenging task is autonomous driving, which demands thorough cognitive processing before decisions can be made.
In this domain, a sequential and interpretive understanding of visual cues is essential for effective perception, prediction, and planning.
Nevertheless, common VQA benchmarks often focus on the accuracy of the final answer while overlooking the reasoning process that enables the generation of accurate responses.
Moreover, existing methods lack a comprehensive framework for evaluating step-by-step reasoning in realistic driving scenarios. To address this gap, we propose DriveLMM-o1, a new dataset and benchmark specifically designed to advance step-wise visual reasoning for autonomous driving. 
Our benchmark features over 18k VQA examples in the training set and more than 4k in the test set, covering diverse questions on perception, prediction, and planning, each enriched with step-by-step reasoning to ensure logical inference in autonomous driving scenarios.
We further introduce a large multimodal model that is fine-tuned on our reasoning dataset, demonstrating robust performance in complex driving scenarios. In addition, we benchmark various open-source and closed-source methods on our proposed dataset, systematically comparing their reasoning capabilities for autonomous driving tasks. Our model achieves a +7.49\% gain in final answer accuracy, along with a 3.62\% improvement in reasoning score over the previous best open-source model. Our framework, dataset, and model are available at \href{https://github.com/ayesha-ishaq/DriveLMM-o1}{https://github.com/ayesha-ishaq/DriveLMM-o1}.
\end{abstract}       
\begin{figure*}[ht]
    \centering
    \includegraphics[width=\linewidth]{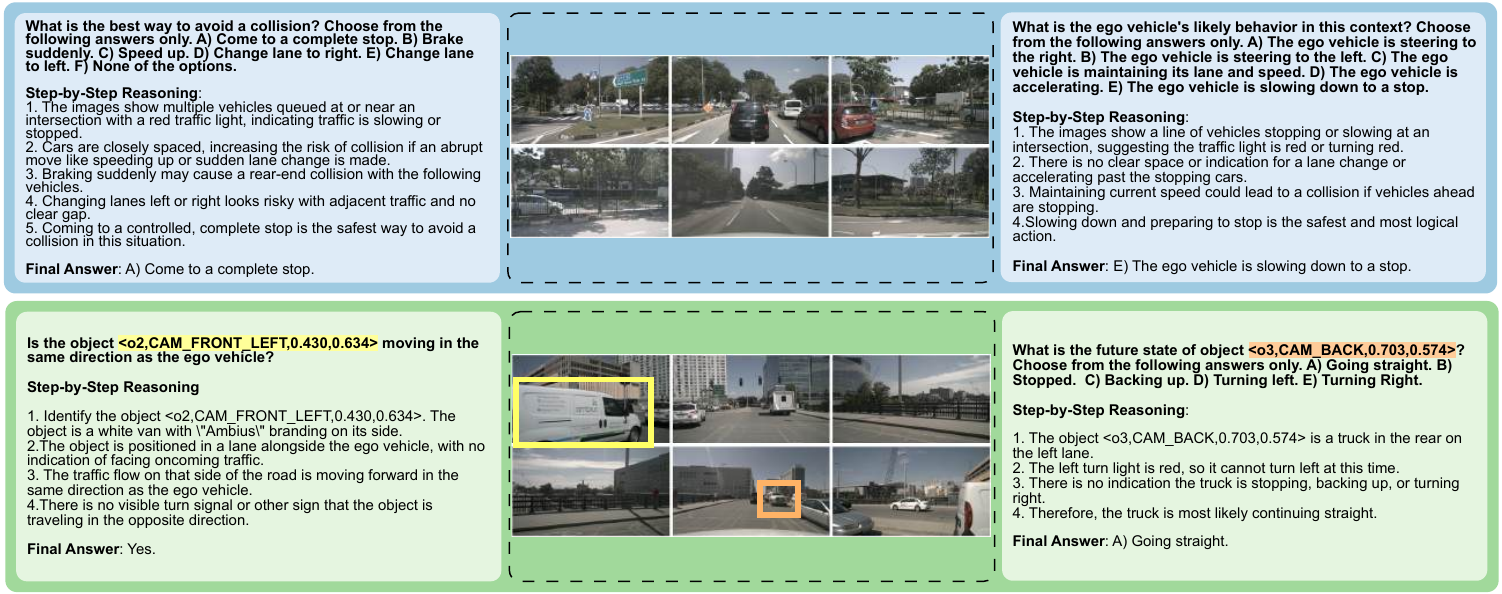}
    \caption{\textbf{Examples from our proposed DriveLMM-o1 dataset.} Our proposed dataset is designed to promote step-by-step reasoning in autonomous driving scenarios, guiding models from understanding the driving task and scene context to making logical inferences based on visual and spatial cues, ultimately leading to accurate decision-making and answer generation. The middle section shows multiview images for two driving scenarios. In the top row, two examples of multiple-choice questions are displayed, along with the step-by-step reasoning annotation required to reach an accurate conclusion. The bottom row includes questions referring to key objects in the scene, with each object highlighted in its corresponding color.}
    \label{fig:examplesl}
\end{figure*}
\section{Introduction}

In autonomous driving contexts, visual reasoning is vital in understanding road environments, interpreting dynamic interactions, and making informed decisions in real time. 
The ability to reason through complex, multi-step tasks is fundamental for autonomous driving, where accurate perception, prediction, and planning are essential for safe and effective navigation. 
Attempts to integrate language models into autonomous driving for Visual Question Answering (VQA) have primarily focused on final-answers, where models generate direct responses based on driving scene inputs. Large Multimodal Models (LMMs) have been applied to perception, such as identifying relevant objects, prediction of potential risks or traffic flow, and planning optimal driving actions, often leveraging large scale datasets to optimize accuracy \cite{drivelm, drivegpt4, drivingwithllms, nuinstruct, Qian_Chen_Zhuo_Jiao_Jiang_2024}.
Despite recent advancements, existing approaches often emphasize end-task accuracy rather than the quality of intermediate reasoning steps. This focus limits the ability to assess and improve the models' logical consistency and explainability in real-world driving scenarios.

Nonetheless, multiple efforts to integrate step-by-step reasoning have been introduced in other domains,
such as natural language processing \cite{2020-commonsense}, mathematical problem-solving \cite{mathsolving}, and scientific reasoning \cite{lu2022learn}. Techniques like Supervised finetuning (SFT) of Large Language Models (LLMs) \cite{rajani2019explain}, Chain-of-Thought prompting \cite{wei2022chain}, and Reinforcement Learning \cite{hurst2024gpt} to generate detailed rationales have improved performance in complex tasks by breaking them down into smaller, sequential steps. Although these efforts have demonstrated improvements in reasoning capabilities across various domains, there remains a gap in applying these principles in high-stakes applications such as autonomous driving, where reasoning must integrate visual and contextual information to achieve accurate and interpretable outcomes.

The effectiveness of reasoning in autonomous driving models heavily depends on the quality of the datasets used for training and learning mechanisms that support the development of step-by-step reasoning processes. 
Previous studies have shown that by prompting or finetuning LMMs to produce step-by-step explanations, significant improvements can be achieved in reasoning tasks \cite{lampinen2022tell,wangself,wei2022chain,thawakar2025llamav}.
Similarly, such structured learning mechanisms that support multi-step reasoning would enable autonomous driving models to make more interpretable and logically consistent decisions.

To this end, our work introduces DriveLMM-o1, an extensive dataset and benchmark designed to evaluate the reasoning capabilities of models in autonomous driving. 
We introduce a diverse set of VQAs that emphasize step-by-step reasoning. 
Additionally, our dataset incorporates multiview images and LiDAR point clouds to enable a comprehensive scene understanding. 
Figure \ref{fig:examplesl} shows a few examples from our proposed benchmark. 
To ensure a thorough assessment, we introduce driving-specific evaluation metrics that gauge the logical coherence and accuracy of model-generated explanations. 
Furthermore, we propose an LMM that is finetuned on our dataset,
demonstrating how learned reasoning enhances final answer quality and contributes to more reliable autonomous driving decisions.\\

In summary, our contributions are as follows:
\begin{itemize}
    \item We introduce a dataset and benchmark specifically designed to assess the reasoning capabilities of models in autonomous driving, capturing diverse real-world scenarios.
    \item We annotate driving scenes that inherently contain rich inputs, including multiview images, LiDAR point clouds, and temporal information, facilitating the integration of various modalities for future VQA solutions.
    \item We propose a novel evaluation metric tailored to autonomous driving, measuring the logical coherence and accuracy of model-generated explanations.
    \item We evaluate previous open-source and closed-source models on our proposed benchmark and introduce a model trained on our step-by-step reasoning dataset. Experimental results show that our model outperforms all models in both reasoning score and final accuracy.
\end{itemize}
\section{Related Work}
In this section, we present methods related to Visual Q Question Answering (VQA) in the context of autonomous driving and general visual reasoning, highlighting their contributions and limitations in relation to our study.

\subsection{Visual Reasoning}
Large Language Models (LMMs) have recently demonstrated enhanced reasoning abilities through various techniques, including Chain-of-Thought (CoT) prompting \cite{wei2022chain} and reinforcement learning \cite{rafailov2023direct}. Visual reasoning extends this capability by requiring models to analyze complex visual data and construct logical step-by-step interpretations. Visual CoT \cite{chen2024visual} enhances knowledge-based visual reasoning by integrating visual perception with language-based reasoning through see, think, and confirm stages. This iterative process grounds visual concepts, adapts them into textual prompts, and verifies reasoning, ensuring improved accuracy, transparency, and efficiency. LLaVA-CoT \cite{xu2024llava} introduces a novel approach to visual reasoning by enabling autonomous multi-stage reasoning, incorporating sequential steps of summarization, visual interpretation, logical analysis, and conclusion generation, significantly improving structured reasoning in large multimodal models. 
Another work demonstrates that training LMMs on short answers limits their ability to generalize to reasoning tasks that demand more detailed explanations \cite{zhang2024improve}.
LlamaV-o1 \cite{thawakar2025llamav} introduces VRC-Bench, the first benchmark for multimodal step-by-step visual reasoning across diverse domains, and proposes a multimodal reasoning model trained with curriculum learning and optimized with beam search, achieving enhanced accuracy. These works demonstrate the value of structured visual reasoning, emphasizing the need for a similar approach in autonomous driving.

\begin{table*}[t]
\centering
\renewcommand{\arraystretch}{1.1} 
\setlength{\tabcolsep}{2pt} 
\caption{\textbf{Comparison of datasets and benchmark statistics in autonomous driving QA research.} The table provides an overview of existing benchmarks, highlighting key attributes such as the number of frames and QA pairs, availability of step-by-step reasoning, input modalities, the number of image views, annotation methods, and data sources. Our benchmark stands out by integrating multiview images and LiDAR point clouds with manually curated step-by-step reasoning. Dataset here reports only the train set statistics.}
\begin{tabularx}{\linewidth}{>{\raggedright\arraybackslash}p{2.7cm} 
                                >{\centering\arraybackslash}p{1cm} 
                                >{\centering\arraybackslash}p{1cm} 
                                >{\centering\arraybackslash}p{1cm} 
                                >{\centering\arraybackslash}p{1cm} 
                                >{\centering\arraybackslash}p{1.5cm}
                                >{\centering\arraybackslash}p{2cm}
                                >{\centering\arraybackslash}p{1cm}
                                >{\raggedright\arraybackslash}p{2.2cm}
                                >{\raggedright\arraybackslash}p{3.1cm}} 
\toprule

&\multicolumn{2}{c}{\textbf{Dataset}} & \multicolumn{7}{c}{\textbf{Benchmark}} \\ 
\cmidrule(lr){2-3} \cmidrule(lr){4-10}
\textbf{} & \textbf{Frames} & \textbf{QAs} & \textbf{Frames} & \textbf{QAs} &  \textbf{Step-by-step Reasoning}& \textbf{Input Modalities}& \textbf{Image Views}&\textbf{Final Annotations}& \textbf{Source} \\ 
\midrule
BDD-X \cite{bddx}         &  5,588    & 23k   & 698  & 2652   & \xmark & Video& 1&Manual& Berkeley Deep Drive  \\ 
NuScenes-QA \cite{Qian_Chen_Zhuo_Jiao_Jiang_2024}   & 28k    & 376k   & 6019  & 83k   & \xmark & Images, Points& 6& Automated& NuScenes \\ 
DriveLM \cite{drivelm}         & 4063  & 377k   & 799   & 15k   & \xmark & Images & 6 & Mostly-Automated& NuScenes, CARLA\\ 
LingoQA \cite{lingoQA}         & 28k    & 420k   & 100   & 1000        & \xmark & Video& 1& Manual& -\\
Reason2Drive \cite{nie2024reason2drive}    & 420k    & 420k   & 180k   & 180k        & \xmark & Video& 1 &Automated&NuScenes, Waymo, OPEN\\
\midrule
DrivingVQA \cite{drivingvqa}      & 3142     & 3142      & 789  & 789   & \checkmark & Image& 2 &Manual& Code de
la Route\\ 
\textbf{DriveLMM-o1 (Ours)}    & 1962   & 18k    & 539   & 4633 & \checkmark & Images, Points& 6 &Manual& NuScenes\\ 
\bottomrule
\end{tabularx}
\label{tab:dataset-comparison}
\end{table*}

\subsection{Visual Question Answering in Autonomous Driving}
VQA in autonomous driving has gained increasing attention as researchers seek to develop models capable of analyzing complex driving scenarios. Several benchmarks and datasets have been introduced to evaluate inference abilities in driving environments. 
NuScenes-QA \cite{Qian_Chen_Zhuo_Jiao_Jiang_2024} extends this effort by offering a multi-modal VQA benchmark, incorporating diverse scenes through multi-view images, point clouds, and question-answer pairs.  LingoQA \cite{lingoQA} introduces a benchmark and dataset for autonomous driving VQA, emphasizing free-form questions beyond perception, and a baseline that identifies an effective Vision Language Model (VLM) setup for late video fusion. Explaining Autonomous Driving Actions with VQA \cite{explainvqa} presents the first empirical study on using VQA to justify autonomous driving decisions and introduces a dataset of image-question-answer triplets for action rationalization. Reason2Drive \cite{nie2024reason2drive} introduces a large-scale benchmark with structured object-centric annotations, focusing on reasoning-specific questions, which make up 27\% of the dataset. Unlike detailed step-by-step reasoning annotations, these questions challenge models to infer logical connections within driving scenarios. The benchmark also introduces ADRScore for evaluating reasoning and proposes a method to integrate perception and prediction into VLMs.
Additionally, DrivingVQA \cite{drivingvqa} introduces a benchmark derived from driving theory tests, emphasizing CoT reasoning to evaluate an agent's ability to reason step-by-step in driving-related decision-making. 


These works have significantly contributed to VQA for autonomous driving, introducing new datasets, benchmarks, and models. However, existing approaches often focus on end-task performance, such as final answer accuracy, while neglecting the quality and transparency of intermediate reasoning steps. Many models treat reasoning as a black-box process, failing to provide structured justifications for their conclusions, which is crucial for safety-critical applications like autonomous driving. Furthermore, existing benchmarks lack multimodal diversity, with many relying primarily on single-view images and limited annotations, making them insufficient for evaluating complex spatial relationships and sequential decision-making.

To address these gaps, our work presents a benchmark and dataset that integrates multimodal inputs, including multiview images and LiDAR point clouds, along with explicit reasoning steps. Additionally, we propose an evaluation framework that assesses the logical coherence of intermediate reasoning and the accuracy of final answers, ensuring a more transparent and reliable assessment of a model’s decision-making process in autonomous driving scenarios.

\section{Autonomous Driving Reasoning Benchmark}

We present DriveLMM-o1, a step-by-step visual reasoning benchmark to comprehensively evaluate reasoning abilities in intricate autonomous driving scenarios for perception, prediction, and planning tasks. This benchmark is designed as a systematic approach to assess the logical explanations and the accuracy of the final predictions generated by models. In Table \ref{tab:dataset-comparison}, we present statistical comparisons of our proposed dataset and benchmark with existing ones in autonomous driving VQA. 
We highlight key attributes, including the number of frames, QA pairs, input modalities, and the availability of step-by-step reasoning. Most existing datasets either lack step-by-step reasoning, have limited dataset sizes, or are constrained in terms of input modalities, such as the absence of multiview images or LiDAR point clouds.
Our benchmark distinguishes itself by incorporating manually curated step-by-step reasoning, providing a more structured and detailed approach to evaluating reasoning capabilities.

\subsection{Benchmark Development} 
\begin{figure}[t]
    \centering
    \includegraphics[width=\linewidth]{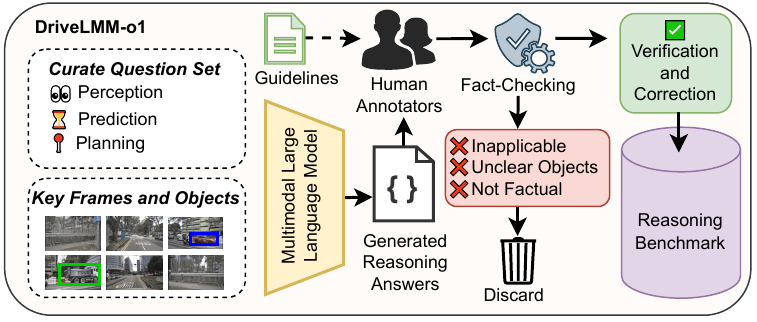}
    \caption{An overview of the benchmark development process. We build our dataset upon frames and objects from NuScenes \cite{nuscenes2019}. Initial reasoning and answers for a standard question set are generated using an LMM, followed by correction and verification of each sample by human annotators.}
    \label{fig:pipeline}
\end{figure}

Figure \ref{fig:pipeline} shows an overview of the process followed to create our proposed benchmark. 
We first generate initial responses utilizing an existing LMM, followed by manual correction and verification of each sample to ensure accuracy and consistency.

\subsubsection*{Question Set Curation}
We carefully design a standardized set of ten questions to ensure comprehensive coverage of VQA tasks across perception, prediction, and planning. These questions address different aspects of autonomous driving scenarios and require a step-by-step logical reasoning process to derive the correct answer, reflecting the decision-making steps essential for safe navigation in autonomous driving scenarios. We extract keyframes and relevant objects from the NuScenes \cite{nuscenes2019} dataset to ground these questions in real-world driving data. The ten-question template is applied to keyframes and objects, generating diverse questions. This approach ensures that the dataset remains well-balanced.

\subsubsection*{Automated Generation}
Once the question set is finalized, we automate the process of generating initial reasoning steps and answers using GPT-4o \cite{hurst2024gpt}. A carefully crafted prompt with explicit instructions and a well-defined output format is provided to guide the model in producing structured reasoning chains. The generated answers follow a logical progression, detailing the observations, key entity relationships, and final outcomes. This automated step ensures efficiency in dataset creation while maintaining coherence in generated responses.

\subsubsection*{Manual Correction and Verification}
Given the critical nature of autonomous driving and the necessity for highly accurate reasoning, we implement a rigorous manual verification process. A team of annotators carefully reviews each generated sample to ensure logical consistency and factual correctness in its reasoning steps and final answer. Questions that do not align with a particular driving scenario and irrelevant objects that do not contribute to meaningful reasoning are discarded. For retained samples, corrections are applied to enhance the completeness of reasoning steps, improve logical flow, and ensure the final answer accurately reflects the driving context. This meticulous verification process guarantees the dataset maintains high-quality reasoning chains, making it a reliable benchmark for evaluating LMMs in autonomous driving scenarios.

\subsection{Evaluation Methodology}

Previous works on evaluating reasoning outputs have explored various approaches. ROSCOE \cite{golovneva2022roscoe} assesses reasoning consistency across multiple metrics, ReCeval \cite{prasad2023receval} focuses on evaluating reasoning correctness through structured rubrics. Both approaches rely on a reference-free method. Alternatively, ReasonEval \cite{xia2024evaluating} examines logical coherence but is specific to mathematical reasoning tasks. These methods are limited in assessing reasoning for autonomous driving VQA tasks, as they do not account for real-world scene understanding or multimodal reasoning.

Recently, VRC-Bench \cite{thawakar2025llamav} introduced a novel evaluation incorporating multimodal understanding and contextual alignment. Following the same strategy, we borrow the Faithfulness-Step, Informativeness-Step, Repetition-Token, Hallucination, Semantic Coverage-Step, Commonsense, and Missing Step metrics from VCR-Bench while we design driving-specific metrics as an extension to access the reasoning steps in light of safety-critical considerations. Table \ref{tab:metrics} shows the details of these metrics. We use GPT-4o \cite{hurst2024gpt} to score the generated steps and answers with reference to the ground truth using a structured evaluation prompt that includes a detailed scoring rubric, predefined rating thresholds, and a standardized JSON output format for consistency. An overall score averages over all the metrics to report a final reasoning score. Additionally, we compute the accuracy of the final answer for all multiple choice question (MCQ) based questions.


\begin{table}[t]
\centering
\renewcommand{\arraystretch}{1} 
\setlength{\tabcolsep}{2pt} 
\caption{Driving-specific attributes assessed in the reasoning steps, scored using ground truth steps as reference.}
\begin{tabularx}{\linewidth}{>{\raggedright\arraybackslash}p{2.5cm} >{\raggedright\arraybackslash}p{5.75cm}} 
\toprule
\textbf{Metric} & \textbf{Detail} \\ 
\midrule
\textbf{Risk Assessment Accuracy} & Evaluates if the model correctly prioritizes high-risk objects or scenarios. \\ 
\textbf{Traffic Rule Adherence} & Scores how well the response follows traffic laws and driving best practices. \\ 
\textbf{Scene Awareness and Object Understanding} & Measures how well the response interprets objects, their positions, and actions. \\ 
\textbf{Relevance} & Measures how well the response is specific to the given scenario and ground truth. \\ 
\textbf{Missing Details} & Evaluates the extent to which critical information is missing from the response. \\ 
\bottomrule
\end{tabularx}
\label{tab:metrics}
\end{table}

\section{Proposed Model}
Our proposed model leverages a large multimodal architecture fine-tuned on our dataset to improve reasoning in autonomous driving scenarios. By integrating stitched multiview images with stepwise reasoning, the model logically infers perception, prediction, and planning tasks before generating a final decision.
\subsubsection*{Multiview Representation for Scene Understanding} Our model processes stitched multiview images to provide a comprehensive view of the driving environment. By combining multiple camera perspectives, such as front, side, and rear views, into a single image, this representation offers a broader spatial context within a single frame. This design choice allows the model to consider objects from multiple viewpoints simultaneously, potentially capturing important spatial relationships without requiring separate image processing for each view. Additionally, consolidating multiple images reduces the need for independent feature extraction across views, leading to improved computational efficiency.

\begin{table*}[t]
\centering
\caption{Performance comparison of close-source and open-source models on our proposed benchmark. We only report the driving-specific metric results here, along with the Overall Reasoning score and Final Answer Accuracy. Our model performs better than other open-source models, improving final answers and reasoning ability.}
\resizebox{\textwidth}{!}{%
\begin{tabular}{lccccc|c|c}
\hline
\textbf{Model} &
  \textbf{\begin{tabular}[c]{@{}c@{}}Risk Assessment\\  Accuracy\end{tabular}} &
  \textbf{\begin{tabular}[c]{@{}c@{}}Traffic Rule\\  Adherence\end{tabular}} &
  \textbf{\begin{tabular}[c]{@{}c@{}}Scene Awareness \\ and Object\\ Understanding\end{tabular}} &
  \textbf{Relevance} &
  \textbf{Missing Details} &
  \textbf{\begin{tabular}[c]{@{}c@{}}Overall\\ Reasoning\end{tabular}} &
  \textbf{\begin{tabular}[c]{@{}c@{}}Final Answer\\ Accuracy\end{tabular}} \\ \hline
\textbf{Closed Source}      &       &       &       &       &       &       &       \\
GPT-4o \cite{hurst2024gpt}                    &  71.32     &   80.72    &  72.96     &   76.65    &   71.43    &   72.52    &    57.84   \\
\hline
\textbf{Open Source}        &       &       &       &       &       &       &       \\
Qwen-2.5-VL-7B \cite{Qwen2VL}             & 46.44 & 60.45 & 51.02 & 50.15 & 52.19 & 51.77 & 37.81 \\
Ovis1.5-Gemma2-9B \cite{lu2024ovis}          & 51.34 & 66.36 & 54.74 & 55.72 & 55.74 & 55.62 & 48.85 \\
Mulberry-7B \cite{yao2024mulberry}                 & 51.89 & 63.66 & 56.68 & 57.27 & 57.45 & 57.65 & 52.86 \\
LLaVA-CoT \cite{xu2024llava}                  & 57.62 & 69.01 & 60.84 & 62.72 & 60.67 & 61.41 & 49.27 \\
LlamaV-o1 \cite{thawakar2025llamav}                 &  60.20     &  73.52     &    62.67   &  64.66     &   63.41    &   63.13    &  50.02     \\
InternVL2.5-8B \cite{chen2024internvl}             & 69.02 & 78.43 & 71.52 & 75.80 & 70.54 & 71.62 & 54.87 \\ \hline
\rowcolor[HTML]{ECF4FF} 
\textbf{DriveLMM-o1 (Ours)} & 73.01 & 81.56 & 75.39 & 79.42 & 74.49 & 75.24 & 62.36 \\  \hline
\end{tabular}%
}
\label{tab:benchmark}
\end{table*}

\subsubsection*{Large Multimodal Model Finetuning} 

To tackle reasoning-based questions and enhance the model's reasoning capabilities and final accuracy, we fine-tune InternVL2.5-8B \cite{chen2024internvl} on our dataset. InternVL2.5 consists of a Vision Transformer (ViT) image encoder and a LLaMA 2-based language model, enabling strong multimodal understanding. The ViT encoder extracts spatial and semantic representations from stitched multiview driving images while the language model processes textual questions.
A key feature of InternVL2.5 is dynamic image patching, which enables variable-resolution image processing by dividing the input image into tiles. This allows the model to handle higher-resolution images efficiently by processing each tile separately and then aggregating their embeddings before passing them to the LLaMA backbone. This dynamic approach ensures that finer details in complex scenes, such as distant objects, road signs, or small hazards, are better captured, which is essential for autonomous driving scenarios.

For finetuning, we freeze both the vision encoder and most of the language model's layers, ensuring the model retains its general multimodal knowledge while adapting to domain-specific reasoning. We use LoRA \cite{hu2022lora} (Low-Rank Adaptation) finetuning in the attention layers of the LLaMA backbone, injecting trainable low-rank matrices into the query, key, and value projections within self-attention. This allows for efficient adaptation with minimal additional parameters, reducing memory requirements while preserving generalizability.

Through this finetuning strategy, the model learns to decompose complex driving scenarios into structured reasoning steps, ensuring a logical progression before arriving at a final decision. By leveraging LoRA and dynamic image patching, our fine-tuned InternVL2.5 model effectively integrates spatial and textual reasoning, refining its ability to process diverse real-world driving situations while maintaining computational efficiency.

\section{Experiments}

In this section, we describe our training data and experimental setup and evaluate the effectiveness of our proposed model. We benchmark it against various closed and open-source models, followed by both quantitative and qualitative analyses.
\subsection{Training Dataset}
The training dataset is constructed following the same methodology as the benchmark but without undergoing rigorous verification. It consists of images from 1,962 driving scenes, covering a diverse range of urban and highway environments. The dataset includes a total of 18,507 reasoning-based Q\&A pairs, designed to assess the model's ability to understand perception, prediction, and planning tasks. Additionally, these Q\&A pairs are accompanied by over 55,000 reasoning steps, ensuring the model learns to generate structured, step-by-step justifications before arriving at a final answer. This extensive dataset enables the model to develop strong reasoning capabilities tailored to autonomous driving scenarios.
\subsection{Experimental Setup}
We use InternVL2.5 8B \cite{chen2024internvl} as our baseline model for our experimental setup. We perform supervised finetuning with a LoRA adapter of rank 16 for one epoch, enabling efficient adaptation while keeping most of the model frozen. Specifically, we freeze the vision encoder and all LLM layers, making only 0.49\% of the total parameters trainable. Our training setup utilizes four NVIDIA A6000 GPUs. To handle high-resolution inputs, we employ dynamic image patching, which allows InternVL to process larger images by dividing them into smaller tiles
, preserving spatial context while accommodating larger field-of-view images.

\subsection{Results and Discussions} In this section, we discuss and compare the quantitative and qualitative benchmarking results. 

\begin{figure*}[t]
    \centering
    \includegraphics[width=0.93\linewidth]{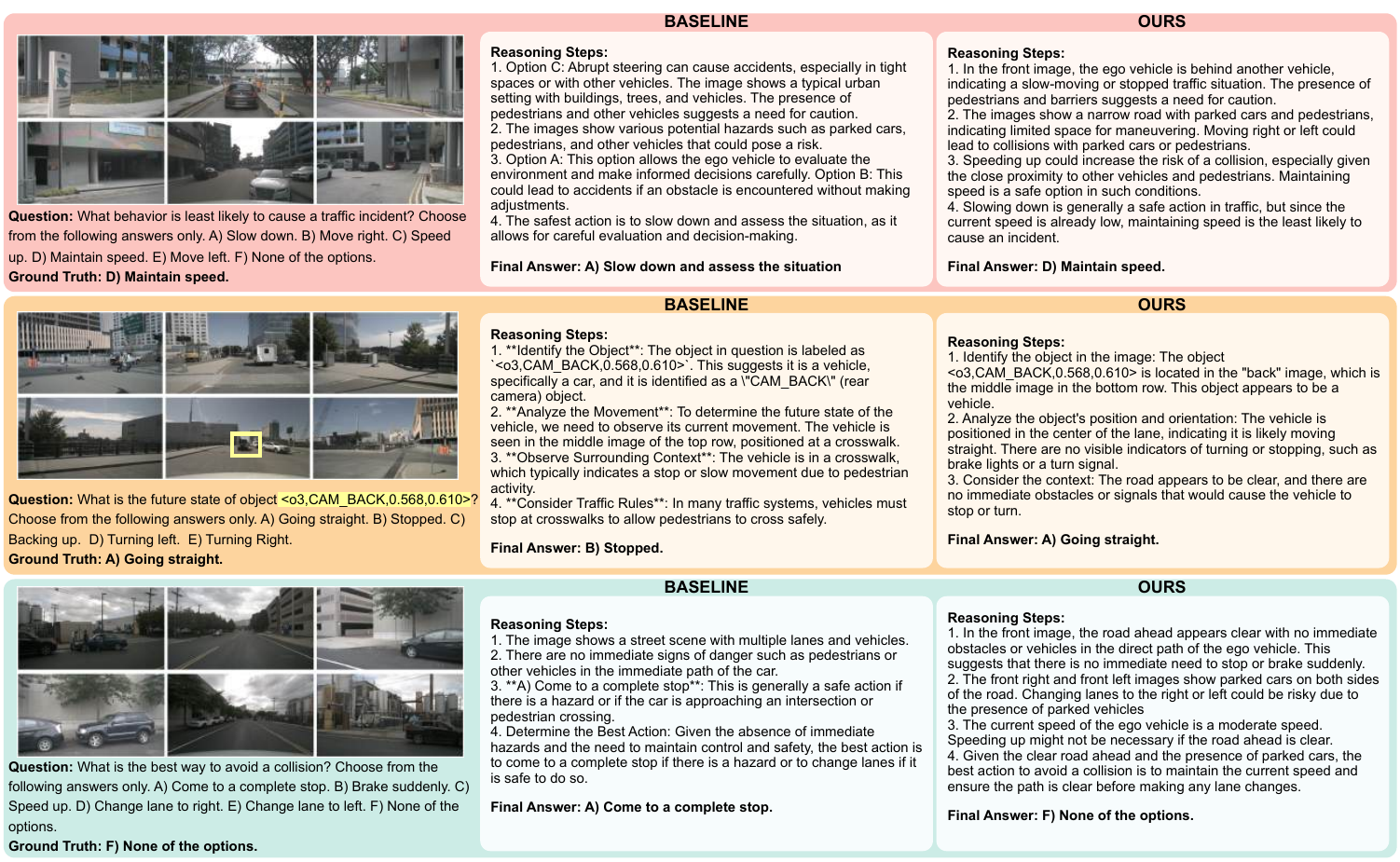}
    \caption{\textbf{Qualitative Results:} We present qualitative examples comparing the reasoning process and final answers generated by the baseline InternVL2.5-8B model and our finetuned model against the ground truth. The results highlight the critical role of accurate reasoning in arriving at the correct final answer, as our model demonstrates greater scene awareness and contextual understanding, leading to more precise and reliable decisions.}
    \label{fig:qualitative}
\end{figure*}
\begin{figure*}[ht]
    \centering
    \includegraphics[width=0.93\linewidth]{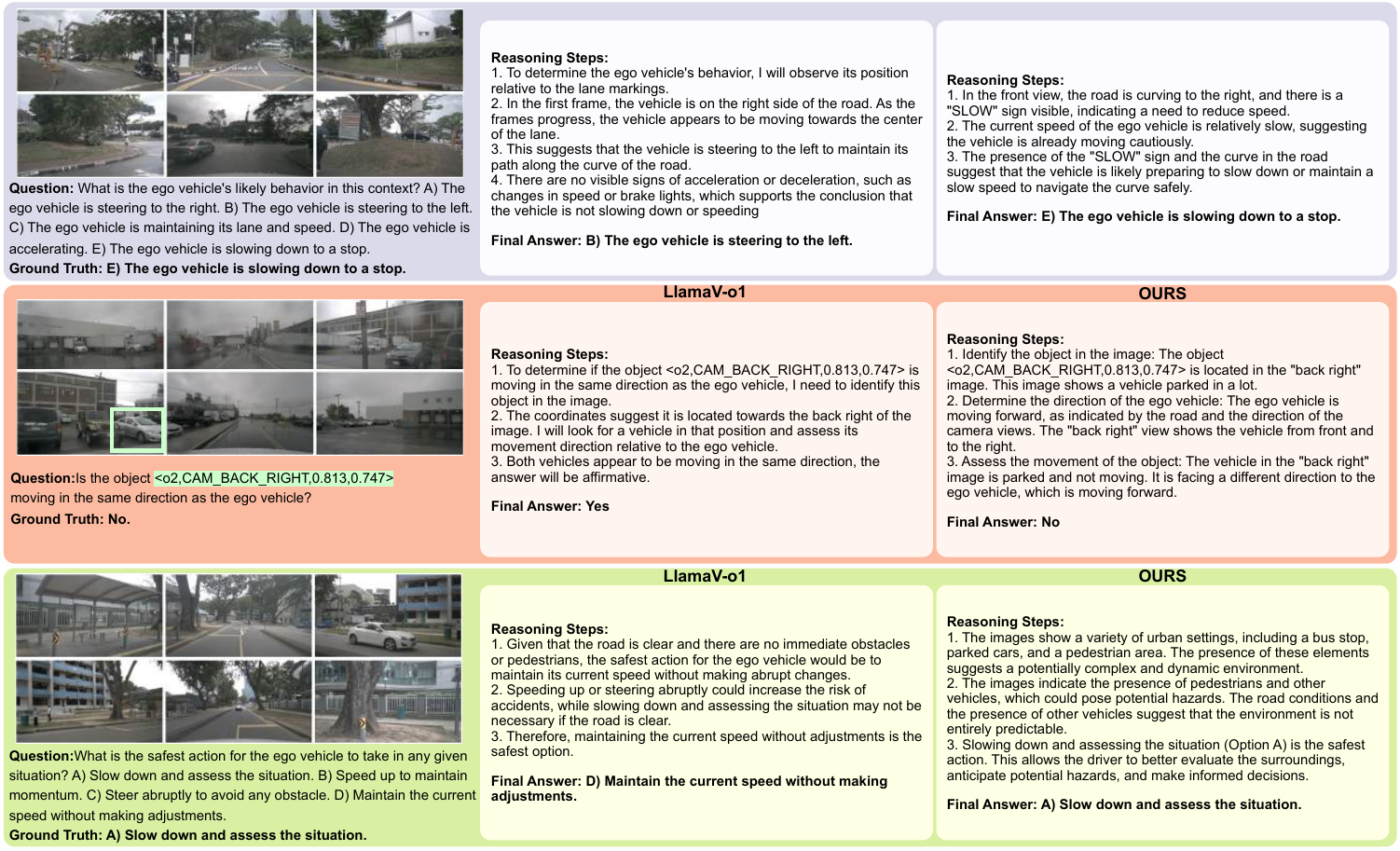}
    \caption{\textbf{Qualitative Comparison against LlamaV-o1:} Qualitative comparison is presented between our model's reasoning outputs and subsequent final answers and the recent Visual Reasoning model LlamaV-o1 \cite{thawakar2025llamav}. 
    While LlamaV-o1 performs well on multiple domains, 
    it struggles with the domain-specific step-by-step logical reasoning required for complex autonomous driving scenarios.
    }
    \label{fig:qualitative_2}
\end{figure*}

\subsubsection*{Quantitative Comaprisons} We evaluate multiple models on our proposed benchmark in a zero-shot setting. The results are reported in Table \ref{tab:benchmark}. Our model, DriveLMM-o1, which is based on InternVL2.5-8B and finetuned on our dataset, outperforms all other open-source models, demonstrating its superior reasoning capabilities in autonomous driving tasks. These evaluations compare model outputs to ground truth annotations, meaning higher scores indicate stronger alignment with human-verified reasoning steps and final decisions.

Across key metrics, DriveLMM-o1 leads in risk assessment accuracy (73.01) and traffic rule adherence (81.56), effectively identifying high-risk elements in driving scenes while making decisions that comply with road regulations. The model also achieves the best scene awareness and object understanding score (75.39), suggesting that it recognizes and understands key objects better than previous models. Additionally, its relevance score (79.42) indicates that responses are well-grounded in the given scenarios, while the missing details score (74.49) confirms that it provides more complete reasoning steps with fewer logical gaps.

Compared to InternVL2.5-8B, the previously best performing open-source baseline, DriveLMM-o1 shows a substantial gain in final answer accuracy (62.36 vs. 54.87), indicating that finetuning on our dataset allows the model to reach more correct conclusions. Moreover, its overall reasoning score (75.24) significantly surpasses other models, reinforcing that DriveLMM-o1 predicts answers and explains its reasoning with higher fidelity.

While models like LlamaV-o1 and LLaVA-CoT perform reasonably well in specific areas, their overall reasoning scores (63.13 and 61.41, respectively) indicate weaker logical consistency in multi-step reasoning for autonomous driving VQA. The zero-shot setting makes this gap more apparent, as these models were not finetuned on our dataset, highlighting the impact of task-specific training.

\subsubsection*{Qualitative Comparison} Figure \ref{fig:qualitative} highlights key differences in reasoning depth and scene specificity between the baseline InternVL2.5-8B and our model.

In the first example, the baseline provides a generic, overly cautious recommendation without fully considering the scene, leading to a broad conclusion. In contrast, our model identifies the ego vehicle’s position, assesses risks, and correctly determines that maintaining speed is the safest action. In the second case, the baseline misidentifies the relevant object and concludes that it is stopped. Our model accurately localizes the object, considers the absence of indicators, and correctly predicts its motion. In the last example, despite clear conditions to proceed, the baseline model advises unnecessary caution, whereas ours makes a more informed decision.

For a thorough analysis, we also present qualitative comparisons between our model, DriveLMM-o1, and the recent LlamaV-o1 \cite{thawakar2025llamav} in Figure \ref{fig:qualitative_2}. LlamaV-o1 showed promising results for reasoning capabilities in multiple visual domains, including medical imaging, complex visual perception, and scientific reasoning. However, it exhibits shortcomings when reasoning for complex autonomous driving questions. In particular, the first and third examples from Figure \ref{fig:qualitative_2} highlight how LlamaV-o1 overlooks safety elements and traffic signs in its reasoning, while our model incorporates these factors, demonstrating a more cautious and context-aware approach. In the second case, LlamaV-o1 fails to understand the spatial setting of the multiview images, whereas our model can determine the views with reference to the ego vehicle.
\section{Conclusion}
In conclusion, we introduce a new dataset and benchmark specifically designed to evaluate reasoning abilities in various real-world scenarios. We also develop a novel evaluation metric that assesses both logical coherence and accuracy in model-generated explanations. By testing a range of existing models on our benchmark and proposing a new model trained on our step-by-step reasoning dataset, we demonstrate that our model surpasses all open-source models in both reasoning scores and final accuracy. These findings highlight the potential for improved reasoning in autonomous driving systems and set the stage for further advancements in the field.






\bibliographystyle{IEEEtran}
\bibliography{IEEEexample}

\begin{thebibliography}{10}
\providecommand{\url}[1]{#1}
\csname url@samestyle\endcsname
\providecommand{\newblock}{\relax}
\providecommand{\bibinfo}[2]{#2}
\providecommand{\BIBentrySTDinterwordspacing}{\spaceskip=0pt\relax}
\providecommand{\BIBentryALTinterwordstretchfactor}{4}
\providecommand{\BIBentryALTinterwordspacing}{\spaceskip=\fontdimen2\font plus
\BIBentryALTinterwordstretchfactor\fontdimen3\font minus \fontdimen4\font\relax}
\providecommand{\BIBforeignlanguage}[2]{{%
\expandafter\ifx\csname l@#1\endcsname\relax
\typeout{** WARNING: IEEEtran.bst: No hyphenation pattern has been}%
\typeout{** loaded for the language `#1'. Using the pattern for}%
\typeout{** the default language instead.}%
\else
\language=\csname l@#1\endcsname
\fi
#2}}
\providecommand{\BIBdecl}{\relax}
\BIBdecl

\bibitem{drivelm}
C.~Sima, K.~Renz, K.~Chitta, L.~Chen, H.~Zhang, C.~Xie, J.~Bei{\ss}wenger, P.~Luo, A.~Geiger, and H.~Li, ``Drivelm: Driving with graph visual question answering,'' in \emph{European Conference on Computer Vision}, 2024.

\bibitem{drivegpt4}
Z.~Xu, Y.~Zhang, E.~Xie, Z.~Zhao, Y.~Guo, K.-Y.~K. Wong, Z.~Li, and H.~Zhao, ``Drivegpt4: Interpretable end-to-end autonomous driving via large language model,'' \emph{IEEE Robotics and Automation Letters}, vol.~9, no.~10, pp. 8186--8193, 2024.

\bibitem{drivingwithllms}
L.~Chen, O.~Sinavski, J.~Hünermann, A.~Karnsund, A.~J. Willmott, D.~Birch, D.~Maund, and J.~Shotton, ``Driving with llms: Fusing object-level vector modality for explainable autonomous driving,'' in \emph{2024 IEEE International Conference on Robotics and Automation (ICRA)}, 2024, pp. 14\,093--14\,100.

\bibitem{nuinstruct}
X.~Ding, J.~Han, H.~Xu, X.~Liang, W.~Zhang, and X.~Li, ``Holistic autonomous driving understanding by bird's-eye-view injected multi-modal large models,'' in \emph{Proceedings of the IEEE/CVF Conference on Computer Vision and Pattern Recognition}, 2024, pp. 13\,668--13\,677.

\bibitem{Qian_Chen_Zhuo_Jiao_Jiang_2024}
\BIBentryALTinterwordspacing
T.~Qian, J.~Chen, L.~Zhuo, Y.~Jiao, and Y.-G. Jiang, ``Nuscenes-qa: A multi-modal visual question answering benchmark for autonomous driving scenario,'' \emph{Proceedings of the AAAI Conference on Artificial Intelligence}, vol.~38, no.~5, pp. 4542--4550, Mar. 2024. [Online]. Available: \url{https://ojs.aaai.org/index.php/AAAI/article/view/28253}
\BIBentrySTDinterwordspacing

\bibitem{2020-commonsense}
\BIBentryALTinterwordspacing
M.~Sap, V.~Shwartz, A.~Bosselut, Y.~Choi, and D.~Roth, ``Commonsense reasoning for natural language processing,'' in \emph{Proceedings of the 58th Annual Meeting of the Association for Computational Linguistics: Tutorial Abstracts}, A.~Savary and Y.~Zhang, Eds.\hskip 1em plus 0.5em minus 0.4em\relax Online: Association for Computational Linguistics, Jul. 2020, pp. 27--33. [Online]. Available: \url{https://aclanthology.org/2020.acl-tutorials.7/}
\BIBentrySTDinterwordspacing

\bibitem{mathsolving}
A.~Lewkowycz, A.~Andreassen, D.~Dohan, E.~Dyer, H.~Michalewski, V.~Ramasesh, A.~Slone, C.~Anil, I.~Schlag, T.~Gutman-Solo \emph{et~al.}, ``Solving quantitative reasoning problems with language models,'' \emph{Advances in Neural Information Processing Systems}, vol.~35, pp. 3843--3857, 2022.

\bibitem{lu2022learn}
P.~Lu, S.~Mishra, T.~Xia, L.~Qiu, K.-W. Chang, S.-C. Zhu, O.~Tafjord, P.~Clark, and A.~Kalyan, ``Learn to explain: Multimodal reasoning via thought chains for science question answering,'' \emph{Advances in Neural Information Processing Systems}, vol.~35, pp. 2507--2521, 2022.

\bibitem{rajani2019explain}
N.~F. Rajani, B.~McCann, C.~Xiong, and R.~Socher, ``Explain yourself! leveraging language models for commonsense reasoning,'' \emph{arXiv preprint arXiv:1906.02361}, 2019.

\bibitem{wei2022chain}
J.~Wei, X.~Wang, D.~Schuurmans, M.~Bosma, F.~Xia, E.~Chi, Q.~V. Le, D.~Zhou \emph{et~al.}, ``Chain-of-thought prompting elicits reasoning in large language models,'' \emph{Advances in neural information processing systems}, vol.~35, pp. 24\,824--24\,837, 2022.

\bibitem{hurst2024gpt}
A.~Hurst, A.~Lerer, A.~P. Goucher, A.~Perelman, A.~Ramesh, A.~Clark, A.~Ostrow, A.~Welihinda, A.~Hayes, A.~Radford \emph{et~al.}, ``Gpt-4o system card,'' \emph{arXiv preprint arXiv:2410.21276}, 2024.

\bibitem{lampinen2022tell}
A.~K. Lampinen, N.~Roy, I.~Dasgupta, S.~C. Chan, A.~Tam, J.~Mcclelland, C.~Yan, A.~Santoro, N.~C. Rabinowitz, J.~Wang \emph{et~al.}, ``Tell me why! explanations support learning relational and causal structure,'' in \emph{International Conference on Machine Learning}.\hskip 1em plus 0.5em minus 0.4em\relax PMLR, 2022, pp. 11\,868--11\,890.

\bibitem{wangself}
X.~Wang, J.~Wei, D.~Schuurmans, Q.~V. Le, E.~H. Chi, S.~Narang, A.~Chowdhery, and D.~Zhou, ``Self-consistency improves chain of thought reasoning in language models,'' in \emph{The Eleventh International Conference on Learning Representations}, 2023.

\bibitem{thawakar2025llamav}
O.~Thawakar, D.~Dissanayake, K.~More, R.~Thawkar, A.~Heakl, N.~Ahsan, Y.~Li, M.~Zumri, J.~Lahoud, R.~M. Anwer \emph{et~al.}, ``Llamav-o1: Rethinking step-by-step visual reasoning in llms,'' \emph{arXiv preprint arXiv:2501.06186}, 2025.

\bibitem{rafailov2023direct}
R.~Rafailov, A.~Sharma, E.~Mitchell, C.~D. Manning, S.~Ermon, and C.~Finn, ``Direct preference optimization: Your language model is secretly a reward model,'' \emph{Advances in Neural Information Processing Systems}, vol.~36, pp. 53\,728--53\,741, 2023.

\bibitem{chen2024visual}
Z.~Chen, Q.~Zhou, Y.~Shen, Y.~Hong, Z.~Sun, D.~Gutfreund, and C.~Gan, ``Visual chain-of-thought prompting for knowledge-based visual reasoning,'' in \emph{Proceedings of the AAAI Conference on Artificial Intelligence}, vol.~38, no.~2, 2024, pp. 1254--1262.

\bibitem{xu2024llava}
G.~Xu, P.~Jin, L.~Hao, Y.~Song, L.~Sun, and L.~Yuan, ``Llava-o1: Let vision language models reason step-by-step,'' \emph{arXiv preprint arXiv:2411.10440}, 2024.

\bibitem{zhang2024improve}
R.~Zhang, B.~Zhang, Y.~Li, H.~Zhang, Z.~Sun, Z.~Gan, Y.~Yang, R.~Pang, and Y.~Yang, ``Improve vision language model chain-of-thought reasoning,'' \emph{arXiv preprint arXiv:2410.16198}, 2024.

\bibitem{bddx}
J.~Kim, A.~Rohrbach, T.~Darrell, J.~Canny, and Z.~Akata, ``Textual explanations for self-driving vehicles,'' in \emph{Proceedings of the European conference on computer vision (ECCV)}, 2018, pp. 563--578.

\bibitem{lingoQA}
\BIBentryALTinterwordspacing
A.-M. Marcu, L.~Chen, J.~Hünermann, A.~Karnsund, B.~Hanotte, P.~Chidananda, S.~Nair, V.~Badrinarayanan, A.~Kendall, J.~Shotton, E.~Arani, and O.~Sinavski, ``Lingoqa: Visual question answering for autonomous driving,'' 2024. [Online]. Available: \url{https://arxiv.org/abs/2312.14115}
\BIBentrySTDinterwordspacing

\bibitem{nie2024reason2drive}
M.~Nie, R.~Peng, C.~Wang, X.~Cai, J.~Han, H.~Xu, and L.~Zhang, ``Reason2drive: Towards interpretable and chain-based reasoning for autonomous driving,'' in \emph{European Conference on Computer Vision}.\hskip 1em plus 0.5em minus 0.4em\relax Springer, 2024, pp. 292--308.

\bibitem{drivingvqa}
C.~Corbi{\`e}re, S.~Roburin, S.~Montariol, A.~Bosselut, and A.~Alahi, ``Drivingvqa: Analyzing visual chain-of-thought reasoning of vision language models in real-world scenarios with driving theory tests,'' \emph{arXiv preprint arXiv:2501.04671}, 2025.

\bibitem{explainvqa}
S.~Atakishiyev, M.~Salameh, H.~Babiker, and R.~Goebel, ``Explaining autonomous driving actions with visual question answering,'' in \emph{2023 IEEE 26th International Conference on Intelligent Transportation Systems (ITSC)}, 2023, pp. 1207--1214.

\bibitem{nuscenes2019}
H.~Caesar, V.~Bankiti, A.~H. Lang, S.~Vora, V.~E. Liong, Q.~Xu, A.~Krishnan, Y.~Pan, G.~Baldan, and O.~Beijbom, ``nuscenes: A multimodal dataset for autonomous driving,'' \emph{arXiv preprint arXiv:1903.11027}, 2019.

\bibitem{golovneva2022roscoe}
O.~Golovneva, M.~Chen, S.~Poff, M.~Corredor, L.~Zettlemoyer, M.~Fazel-Zarandi, and A.~Celikyilmaz, ``Roscoe: A suite of metrics for scoring step-by-step reasoning,'' \emph{arXiv preprint arXiv:2212.07919}, 2022.

\bibitem{prasad2023receval}
A.~Prasad, S.~Saha, X.~Zhou, and M.~Bansal, ``Receval: Evaluating reasoning chains via correctness and informativeness,'' \emph{arXiv preprint arXiv:2304.10703}, 2023.

\bibitem{xia2024evaluating}
S.~Xia, X.~Li, Y.~Liu, T.~Wu, and P.~Liu, ``Evaluating mathematical reasoning beyond accuracy,'' \emph{arXiv preprint arXiv:2404.05692}, 2024.

\bibitem{Qwen2VL}
P.~Wang, S.~Bai, S.~Tan, S.~Wang, Z.~Fan, J.~Bai, K.~Chen, X.~Liu, J.~Wang, W.~Ge, Y.~Fan, K.~Dang, M.~Du, X.~Ren, R.~Men, D.~Liu, C.~Zhou, J.~Zhou, and J.~Lin, ``Qwen2-vl: Enhancing vision-language model's perception of the world at any resolution,'' \emph{arXiv preprint arXiv:2409.12191}, 2024.

\bibitem{lu2024ovis}
S.~Lu, Y.~Li, Q.-G. Chen, Z.~Xu, W.~Luo, K.~Zhang, and H.-J. Ye, ``Ovis: Structural embedding alignment for multimodal large language model,'' \emph{arXiv preprint arXiv:2405.20797}, 2024.

\bibitem{yao2024mulberry}
H.~Yao, J.~Huang, W.~Wu, J.~Zhang, Y.~Wang, S.~Liu, Y.~Wang, Y.~Song, H.~Feng, L.~Shen \emph{et~al.}, ``Mulberry: Empowering mllm with o1-like reasoning and reflection via collective monte carlo tree search,'' \emph{arXiv preprint arXiv:2412.18319}, 2024.

\bibitem{chen2024internvl}
Z.~Chen, J.~Wu, W.~Wang, W.~Su, G.~Chen, S.~Xing, M.~Zhong, Q.~Zhang, X.~Zhu, L.~Lu \emph{et~al.}, ``Internvl: Scaling up vision foundation models and aligning for generic visual-linguistic tasks,'' in \emph{Proceedings of the IEEE/CVF Conference on Computer Vision and Pattern Recognition}, 2024, pp. 24\,185--24\,198.

\bibitem{hu2022lora}
E.~J. Hu, Y.~Shen, P.~Wallis, Z.~Allen-Zhu, Y.~Li, S.~Wang, L.~Wang, W.~Chen \emph{et~al.}, ``Lora: Low-rank adaptation of large language models.'' \emph{ICLR}, vol.~1, no.~2, p.~3, 2022.

\end{thebibliography}

\end{document}